\pdfoutput=1

\documentclass[11pt]{article}

\usepackage[final]{coling}

\usepackage{times}
\usepackage{latexsym}

\usepackage[T1]{fontenc}

\usepackage[utf8]{inputenc}

\usepackage{microtype}

\usepackage{inconsolata}

\usepackage{graphicx}

\usepackage{times}
\usepackage{latexsym}
\usepackage{arydshln}
\usepackage{graphicx}
\usepackage{subcaption}
\usepackage{hyperref}
\usepackage{lscape}
\usepackage{multirow}
\usepackage{booktabs}
\usepackage{amsmath}

\usepackage{algorithm}
\usepackage{algpseudocode}
\usepackage[capitalise]{cleveref} 

\usepackage{microtype,enumitem}
\usepackage{xcolor}	

\usepackage[export]{adjustbox}
\usepackage[utf8]{inputenc}

\usepackage{float}
\newfloat{algorithm}{t}{lop}
\usepackage{adjustbox}

\title{SynDARin: Synthesising Datasets for Automated Reasoning in Low-Resource Languages}

\newcommand\blfootnote[1]{%
  \begingroup
  \renewcommand\thefootnote{}\footnote{#1}%
  \addtocounter{footnote}{-1}%
  \endgroup
}
\makeatother

\author{%
Gayane Ghazaryan$^{\dagger1}$ \hspace{0.5em}
Erik Arakelyan$^{\dagger2}$ \hspace{0.5em} Pasquale Minervini$^{3}$ \hspace{0.5em} \textbf{Isabelle Augenstein}$^2$ \\
$^1$American University of Armenia \quad
$^2$University of Copenhagen \\ $^3$University of Edinburgh \quad \\ \texttt{gayane\_ghazaryan2@edu.aua.am} \
\texttt{erik.a@di.ku.dk} \\ 
\texttt{p.minervini@ed.ac.uk} \
\texttt{augenstein@di.ku.dk}
}

\begin{document}
\maketitle

\blfootnote{$^{\dagger}$Equal contribution}

\begin{abstract}

Question Answering (QA) datasets have been instrumental in developing and evaluating Large Language Model (LLM) capabilities.
However, such datasets are scarce for languages other than English due to the cost and difficulties of collection and manual annotation.
This means that producing novel models and measuring the performance of multilingual LLMs in low-resource languages is challenging. 
To mitigate this, we propose \textbf{S}yn\textbf{DAR}in, a method for generating and validating QA datasets for low-resoucre languages.
We utilize parallel content mining to obtain \emph{human-curated} paragraphs between English and the target language.
We use the English data as context to \emph{generate} synthetic multiple-choice (MC) question-answer pairs, which are automatically translated and further validated for quality.
Combining these with their designated non-English \emph{human-curated} paragraphs form the final QA dataset. 
The method allows to maintain content quality, reduces the likelihood of factual errors, and circumvents the need for costly annotation.
To test the method, we created a QA dataset with $1.2$K samples for the Armenian language.
The human evaluation shows that $98\%$ of the generated English data maintains quality and diversity in the question types and topics, while the translation validation pipeline can filter out $\sim70\%$ of data with poor quality.
We use the dataset to benchmark state-of-the-art LLMs, showing their inability to achieve human accuracy with some model performances closer to random chance.
This shows that the generated dataset is non-trivial and can be used to evaluate reasoning capabilities in low-resource language.

\end{abstract}

\section{Introduction}

\begin{figure*}[t!]
    \centering
    \includegraphics[trim=0.5cm 0.0cm 0.5cm 0.5cm,clip=true,width=0.95\textwidth]{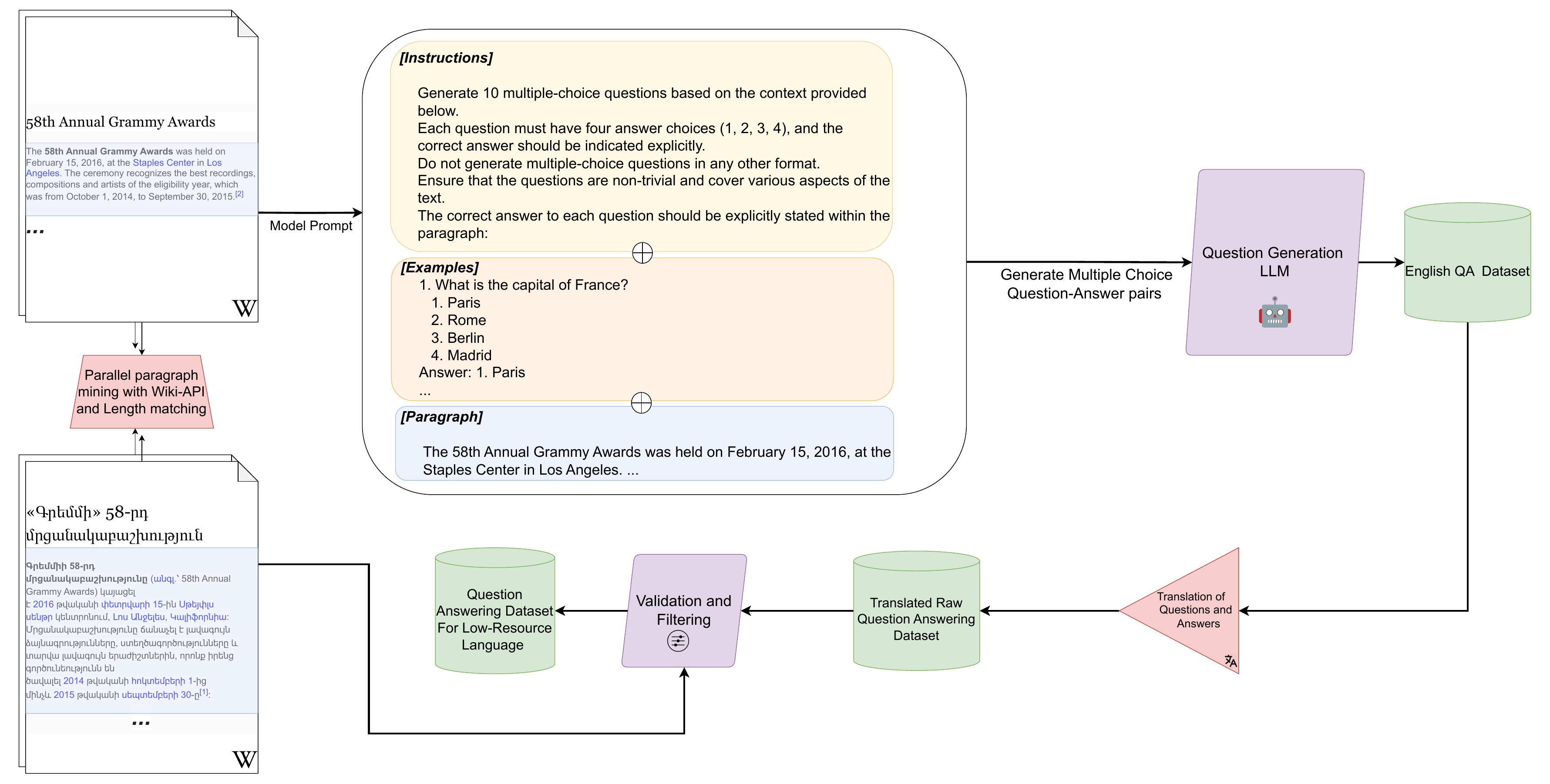}
    \caption{The proposed framework is comprised of three components: (i) a module for mining parallel paragraphs using wiki-API and length matching; (ii) generating a synthetic question-answering dataset with an LLM using the mined English paragraphs; (iii) translating the question-answer pairs and Filtering/Validating them for obtaining a high-quality synthetic QA dataset in the low-resource language.}
    \label{fig:framework}
\end{figure*}

Question Answering (QA) has been a hallmark task for testing reading comprehension and reasoning capabilities in NLP systems.
The availability of numerous English benchmarks that frame the problem as extractive, cloze-style or open-domain \citep{yang2015wikiqa,rajpurkar2016squad,chen2017reading} reasoning tasks, along with novel pre-trained language models (PLMs) \citep{devlin2018bert,lewis2019bart} and LLMs \citep{touvron2023llama,jiang2023mistral,achiam2023gpt} allowed for the development and granular evaluation of QA systems that occasionally boast human-like or better performance \citep{devlin2018bert,min2023recent,rogers2023qa}.
Although some concentrated effort has been made to create multilingual QA resources \citep{lewis2019mlqa,asai2018multilingual,liu2019xqa}, the datasets remain rather scarce and usually cover a small selected set of languages due to the labour-intensive annotation costs. 
The proposed methods suggest using direct machine translation \citep{lewis2019mlqa,carrino2019automatic} or multilingual synthetic data generation \citep{riabi2020synthetic,agrawal2023qameleon,shakeri2020towards}. However, these approaches are directly bound to introduce biases and hallucinations during translation \citep{artetxe2020translation}, cross-lingual transfer \citep{lauscher2020zero, guerreiro2023hallucinations} or generation \citep{ahuja2023mega}. 
%
%
These limitations directly hinder the possibility to \emph{develop} and \emph{evaluate} the multilingual QA capabilities of language models in low-resource languages.

In this work, we propose \textbf{S}yn\textbf{DAR}in, a novel method for synthesising datasets for automated reasoning in low-resource languages that circumvents the above-mentioned obstacles and test it by creating a QA dataset for the Armenian language, which has virtually no presence of structured NLP datasets \citep{avetisyan2023large}.
We mine parallel English and Armenian introductory paragraphs from the same diverse set of Wikipedia articles, ensuring that the contents match by comparing their relative length. Similar mining approaches have been shown to be efficient for this task \citep{lewis2021paq,artetxe2019massively}.
This allows us to obtain human-curated text from diverse topics while bypassing a wide chunk of direct content translation and annotation.
Given the English subset of this data, we generate MC question-answer pairs by prompting an LLM to produce queries with an answer explicitly mentioned within the paragraph.
Following \citet{lewis2019mlqa}, we filter out examples that do not contain the answer substring verbatim in the paragraph and additionally perform a human evaluation on a subset of $50$ examples and show that $98\%$ of these question-answer pairs are answerable and maintain quality.
The produced question-answers are subsequently translated using an automated tool and further validated by answer substring and semantic matching in the parallel Armenian paragraph.
This allows us to mitigate the likelihood of hallucinated, biased and inconsistent entries in the final QA dataset. Our human evaluation with native Armenian speakers shows that ~$70\%$ of such corrupted examples are removed.
We use the dataset as a reasoning benchmark for Armenian and evaluate several LLMs in zero-shot, few-shot, and fine-tuned modes.
We show that the dataset cannot be trivially solved, thus highlighting it as a useful resource for measuring model performance.
In sum, our contributions are as follows: (i) a novel method for QA dataset construction in low-resource languages, (ii) a QA dataset in Armenian, (iii) ablations showing the quality of the generated samples and (iv) an evaluation of several LLM families on the QA dataset.   

\section{Methodology}

An outline of \textbf{S}yn\textbf{DAR}in can be seen in \cref{fig:framework}.

\subsection{Parallel Data Mining}

Given parallel English and Armenian introductory paragraph tokens $\mathcal{P}_{\text{En}} = (T_1, \dots T_n)$, $\mathcal{P}_{\text{Arm}} = (T_1,\dots T_m)$ obtained from a diverse set of Wiki articles, we want to save the segments that contain the same content. As the introductory paragraphs in Wikipedia contain highly similar information \citep{lewis2019mlqa}, we found that filtering out the paragraph pairs based on their relative view count and the number of tokens, i.e. length, is sufficient. To do this, we simply define a conditional rejection process on Wikipedia pages that have been viewed more than $1000$ and edited more than 5 times $| \lVert \mathcal{P}_{\text{En}}\rVert-\lVert\mathcal{P}_{\text{Arm}}\rVert | \leq K_{\text{DM}}$, where $K_{\text{DM}}$ is the threshold for the length difference. A higher length difference would imply that the contents of the paragraphs are misaligned, thus making us reject such samples. Consequently, we are able to obtain naturally written human-curated parallel paragraphs that cover a diverse set of topics.

\subsection{QA Generation}

After obtaining the parallel data, we prompt an LLM $\mathcal{M}$ with instructions $\mathcal{I}=(T_1, \dots T_{|\mathcal{I}|})$ and $10$ in-context example demonstrations $\mathcal{E} = (E_1,\dots E_{10})$, where $\forall i, E_i=(T_1,\dots T_{|E_i|})$, to generate diverse English MC question-answer pairs $\mathcal{K_{\text{Eng}}}=\left\{\left(q_1, a_1\right) \ldots\left(q_N, a_N\right)\right\}$ given an English context paragraph $\mathcal{P}_{\text{En}}$:
\newcommand\shortdots{\makebox[0.75em][c]{.\hfil.\hfil.}}
\small
\begin{align}
    q_i, a_i \sim \prod_{t=1}^{|\mathcal{K}_i|} P_{\mathcal{M}}\left(T_t^{(i)} \mid T_1^{(i)}, \shortdots, T_{t-1}^{(i)}, \mathcal{I}, \mathcal{E}, \mathcal{P}_{\text{En}}\right)
\end{align}
\normalsize

\begin{table}[t!]
\centering
\begin{adjustbox}{max width=\columnwidth}
\begin{tabular}{lccccccccc}
\toprule
  \textbf{Who} & \textbf{Where} & \textbf{What} & \textbf{When} & \textbf{Which} & \textbf{How} & \textbf{General} & \textbf{Why} \\
304 & 128 & 1536 & 215 & 473 & 244 & 76 & 16 \\
\bottomrule
\end{tabular}
\end{adjustbox}
\caption{Frquency of Question Types in the generated English question-answer pairs.}
\label{tab:qa_types}
\end{table}

We filter out all repeating questions, $\forall \{i,j: i\neq j\}, q_i\neq q_j$, and question-answers pairs where the answer span is not exactly mentioned within the text, i.e. $a_i \not\subset  \mathcal{P}_{\text{En}}$. An example input used for generation can be seen in \cref{fig:framework}. This generation and validation pipeline resembles the ones in \citet{lewis2021paq, agrawal2023qameleon}, which have shown successful question-generation results for the English language. Several examples of produced questions are available in \cref{sec:appendix}.

\subsection{Translation and Validation}

We transfer the generated question-answer pairs $\mathcal{K_{\text{Eng}}}$ into Armenian by using the Google Translate API to obtain $\mathcal{K_{\text{Arm}}}$. To mitigate the inconsistencies introduced during the translation process, we save only the samples where the translated answer $a_i \in \mathcal{K_{\text{Arm}}}$ is contained within and semantically related to the paragraph $\mathcal{P}_{\text{Arm}}$. To do this, we use a fuzzy substring matching function $\mathcal{F}: \mathcal{T} \times \mathcal{T} \rightarrow [0,1] $, along with a multilingual language model $\mathcal{M}_{\text{sim}}:\mathcal{T} \rightarrow \mathcal{R}^d$ to measure semantic similarity, where $\mathcal{T}$ is an arbitrary set of tokens and $d$ is the dimensionality of the embedding space of the model. Samples below a certain threshold, $\mathcal{F}(a_i, \mathcal{P}_{\text{Arm}}) \leq K_{\text{Fuzz}} \text{ and } \cos(\mathcal{M}(a_i), \mathcal{M}(\mathcal{P}_{\text{Arm}}))\leq K_{\text{Sim}}$ are filtered out. Note that exact matching is insufficient, as the morphology of the translated answer tokens can vary in the low-resource language. The multiple-choice answers are balanced uniformly in the final dataset so as not to introduce a bias toward any particular answer ordering.

\begin{table}[t!]
\begin{adjustbox}{max width=\columnwidth}
{
\begin{tabular}{@{}lll@{}}
\toprule
\textit{\textbf{Problem type(\%)}} & \textit{Filtered} & \textit{Unfiltered} \\ \midrule
Partially Missing Info & 38 & 77 \\
Bad Translation     & 5  & 51 \\
Partially Correct Answers     & 22 & 31 \\
Several Correct Answers       & 27 & 45 \\
Date Mismatch                 & 13 & 17 \\
Other                         & 8  & 22 \\ \bottomrule
\end{tabular}
}
\end{adjustbox}
\caption{Unanswerable sample analysis before(Unfiltered) and after(Filtered) the validation. Annotators can choose multiple reasons per sample.}
\label{tab:human_annot}
\end{table}

\section{Experimental Setup}

\begin{figure}[t!]
    \centering
    \includegraphics[width=1.1\columnwidth]{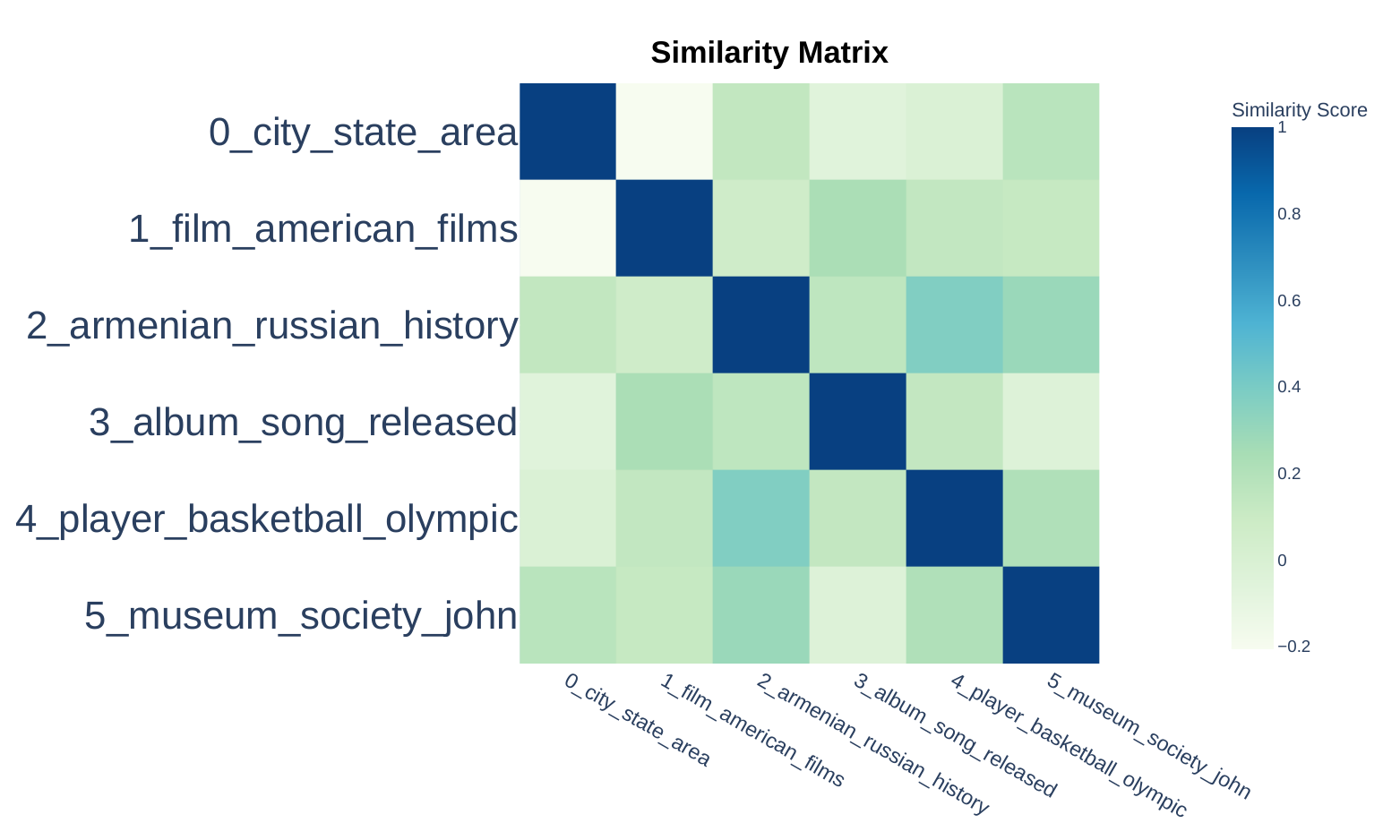}
    \caption{BERTopic embeddings similarity heatmap for the top 6 frequent topics in the mined English paragraphs.}
    \label{fig:topic_dist}
\end{figure}

\paragraph{QA Generation}
Our QA generation uses GPT-4 \cite{achiam2023gpt}, known for generating high-quality text \citep{zhou2023synthetic} and synthetic data \citep{hamalainen2023evaluating,li2023synthetic}.

\paragraph{Substring Matching and Semantic Similarity} 
We employ Levenshtein distance for fuzzy substring matching ($\mathcal{F}$) and multilingual sentence embeddings \citep{reimers-2019-sentence-bert} ($\mathcal{M}_{\text{sim}}$) for semantic similarity using cosine distance.

\paragraph{Armenian QA Benchmarking}
We benchmark GPT-3.5 \cite{achiam2023gpt}, CMD-R, and CMD-R+ \cite{cohere2024commandr} using $\{0,2,4,6\}$ in-context examples with few-shot prompting \cite{brown2020language} on the Armenian QA dataset. We further frame the task as classification with multiple-choice answers and perform supervised fine-tuning with a recipe \citep{mosbach2020stability} on XLM-RoBERTa-base \cite{conneau2019unsupervised}, with $\{32,64, \dots ,980\}$ training samples and benchmark it on the same testing set. Following \citet{poliak2018hypothesis}, we analyze model performance on \emph{question-only} and \emph{paragraph-only} inputs for bias detection.

\section{Results}

\subsection{English QA Dataset Generation}
We mined $300$ parallel English-Armenian Wikipedia paragraphs and generated $10$ diverse questions with $4$ MC answers each, resulting in $3000$ English QA pairs.

\paragraph{Dataset Diversity}
We assessed question diversity (\cref{tab:qa_types}) and found meaningful variation consistent with prior human-curated datasets \citep{lewis2019mlqa, rajpurkar2016squad}. Topic modelling using BERTopic \citep{grootendorst2022bertopic} validated the subject diversity (\cref{fig:topic_dist}). A granular diversity analysis within the dataset is presented in \cref{sec:appendix}.

\paragraph{Human Evaluation}
To assess the data quality, we follow \citet{lewis2021paq} and ask two English-speaking human annotators to manually inspect $50$ randomly chosen samples from the English QA dataset regarding the captured contextual information and answerability of the sample question. The results show, with an inter-annotator agreement score of Cohen's $\kappa = 0.99$, that $98\%$ of examples contain sufficient details to answer the question while accurately capturing contextual information.

\subsection{Automatic Translation and Validation}
We translate the obtained $3000$ QA samples and pass the results through our validation pipeline to produce $1235$ filtered Armenian examples.  

\begin{table}[t!]
\centering
\begin{adjustbox}{max width=\columnwidth}
\begin{tabular}{@{}lcccc@{}}
\toprule
& \multicolumn{4}{c}{Accuracy} \\
\cmidrule{2-5}
Filter & \texttt{128} & \texttt{256} & \texttt{512} & \texttt{987} \\ \midrule
\emph{Complete}  &30.1\%          & 33.5\%       & 38.7\%       & 39.5\%       \\
\emph{paragraph-only} & \underline{26.7\%} & \underline{28.3\%}       & \underline{23.9\%}       & \underline{28.3\%}       \\
\emph{question-only} & \underline{22.1\%}  & \underline{22.7\%}       & \underline{19.4\%}       & \underline{23.5\%}       \\
\emph{Random performance} & \multicolumn{4}{c}{\textbf{25.0\%}} \\ \bottomrule
\end{tabular}
\end{adjustbox}
\caption{The results of fine-tuning XLM-Roberta on the Armenian QA dataset with a varying number of training samples in different degeneracy testing scenarios.}
\label{tab:xlm_roberta_short}
\end{table}

\paragraph{Armenian QA dataset}

We use these samples and their designated Armenian paragraphs to form the QA dataset. We split the data into $80/20$ \emph{train/test} buckets with $987$ samples in training and $247$ in testing. We ensure that the paragraphs in the testing set are not contained in the train set to avoid any data leakage. We maintain a uniform distribution of MC questions within the answers, avoiding bias towards any answer ordering.

\paragraph{Human Evaluation}
We assessed the translation validation pipeline and datasets using two native-speaking annotators. They reviewed the \emph{test} set, which was mixed with 100 randomly flagged poor samples from automatic validation. Annotators either answered the samples or marked them as unanswerable, citing reasons from a predefined set, see in \cref{tab:human_annot}. Results showed that $87\%$ of the flagged examples were unanswerable due to insufficient context, translation errors, or hallucinations. The error breakdown in \cref{tab:human_annot} highlights the quality improvement in filtered samples w.r.t. to the abovementioned discrepancies, where annotators answered correctly in $75\%$ of cases. We measure the inter-annotator agreement using Cohen's $\kappa=0.8$. These confirm the ability of our validation pipeline to maintain the dataset quality. 

\paragraph{Benchmarks}
\begin{table}[t!]
\centering
\begin{adjustbox}{max width=\columnwidth }
\begin{tabular}{@{}lcccc@{}}
\toprule
& \multicolumn{4}{c}{Accuracy} \\
\cmidrule{2-5}
Model Name    & \texttt{0} & \texttt{2} & \texttt{4} & \texttt{6} \\ \midrule
Command-R          & 58.7\%       & \textbf{68.4}\%       & \underline{64.8}\%       & 64.0\%       \\
Command-R+      & 59.3\%       & 67.2\%       & \underline{69.6}\%       & \textbf{70.9}\%       \\
GPT-3.5           & 56.3\%       & \underline{56.3}\%       & \textbf{59.1}\%       & 54.3\%       \\ \bottomrule
\end{tabular}
\end{adjustbox}
\caption{Model Accuracy with a varying number of provided in-context samples before generation.}
\label{tab:model_accuracy}
\end{table}

To show the value of the created dataset, we investigate if it suffers from statistical biases or degenerate solutions by training an XLM-RoBERTa model on inputs that contain only the paragraph or the question, excluding everything else from the sample. The results in \cref{tab:xlm_roberta_short} show that regardless of the number of training samples, the models trained with question and paragraph-only samples behave similarly to random chance, while training with complete data gradually increases the performance, highlighting that the dataset is unlikely to suffer from inconsistencies and degenerate solutions and can be used for developing QA capabilities for Armenian. We further benchmark several state-of-the-art LLMs on this dataset in supervised fine-tuning, \emph{zero-shot} and \emph{few-shot} settings. We see in \cref{tab:model_accuracy} that even the largest models do not trivially solve the dataset, showing its utility as a benchmarking tool. 

\section{Conclusion}

We propose \textbf{S}yn\textbf{DAR}in, a novel method for constructing QA datasets for low-resource languages and produce a dataset for the Armenian language. Systematic studies of the reliability of the individual modules to produce diverse QA samples that maintain answerability and quality show the effectiveness of the method. We further use the produced Armenian QA dataset to benchmark state-of-the-art LLMs and show the value of the proposed resource in evaluating QA reasoning capabilities in the low-resource language.

\section*{Limitations}

The proposed methods have currently been tested only for a smaller-scale QA dataset creation in Armenian, thus not allowing us to complete a wider cross-lingual study. The study benchmarks should be extended and analyzed further in more multilingual, low-resource languages. In the case of extremely rare low-resource languages, the automatic translation part within our pipeline would require either the development of such a translation method, robust cross-lingual transfer from a similar language or direct manual effort, all of which are bound to introduce either qualitative or logistic complications while creating the final QA resource.

\section*{Acknowledgments}
$\begin{array}{l}\includegraphics[width=1cm]{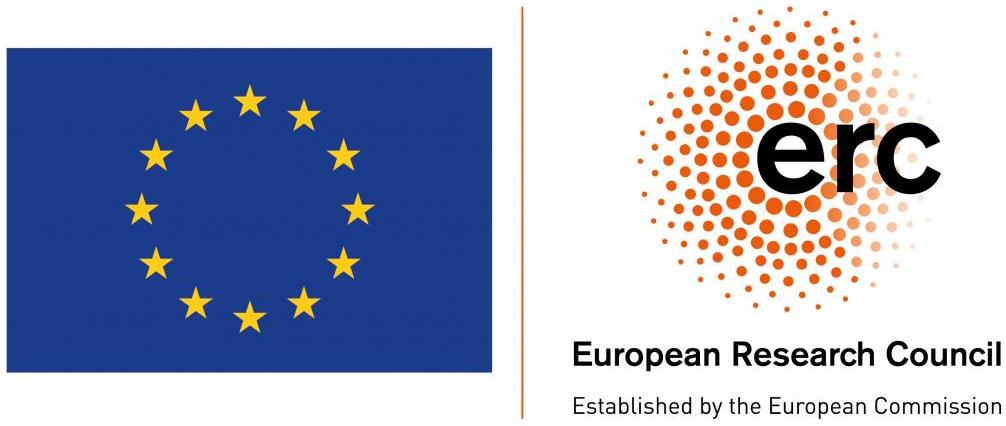} \end{array}$ 
Erik is partially funded by a DFF Sapere Aude research leader grant under grant agreement No 0171-00034B, as well as by a NEC PhD fellowship, and is supported by the Pioneer Centre for AI, DNRF grant number P1.
Pasquale was partially funded by ELIAI (The Edinburgh Laboratory for Integrated Artificial Intelligence), EPSRC (grant no.\ EP/W002876/1), an industry grant from Cisco, and a donation from Accenture LLP.
Isabelle's research is partially funded by the European Union (ERC, ExplainYourself, 101077481), and is supported by the Pioneer Centre for AI, DNRF grant number P1.
This work was supported by the Edinburgh International Data Facility (EIDF) and the Data-Driven Innovation Programme at the University of Edinburgh.
%

\bibliography{custom}

\newpage
\appendix
\begin{figure}[H]
    \centering
    \includegraphics[width=\columnwidth]{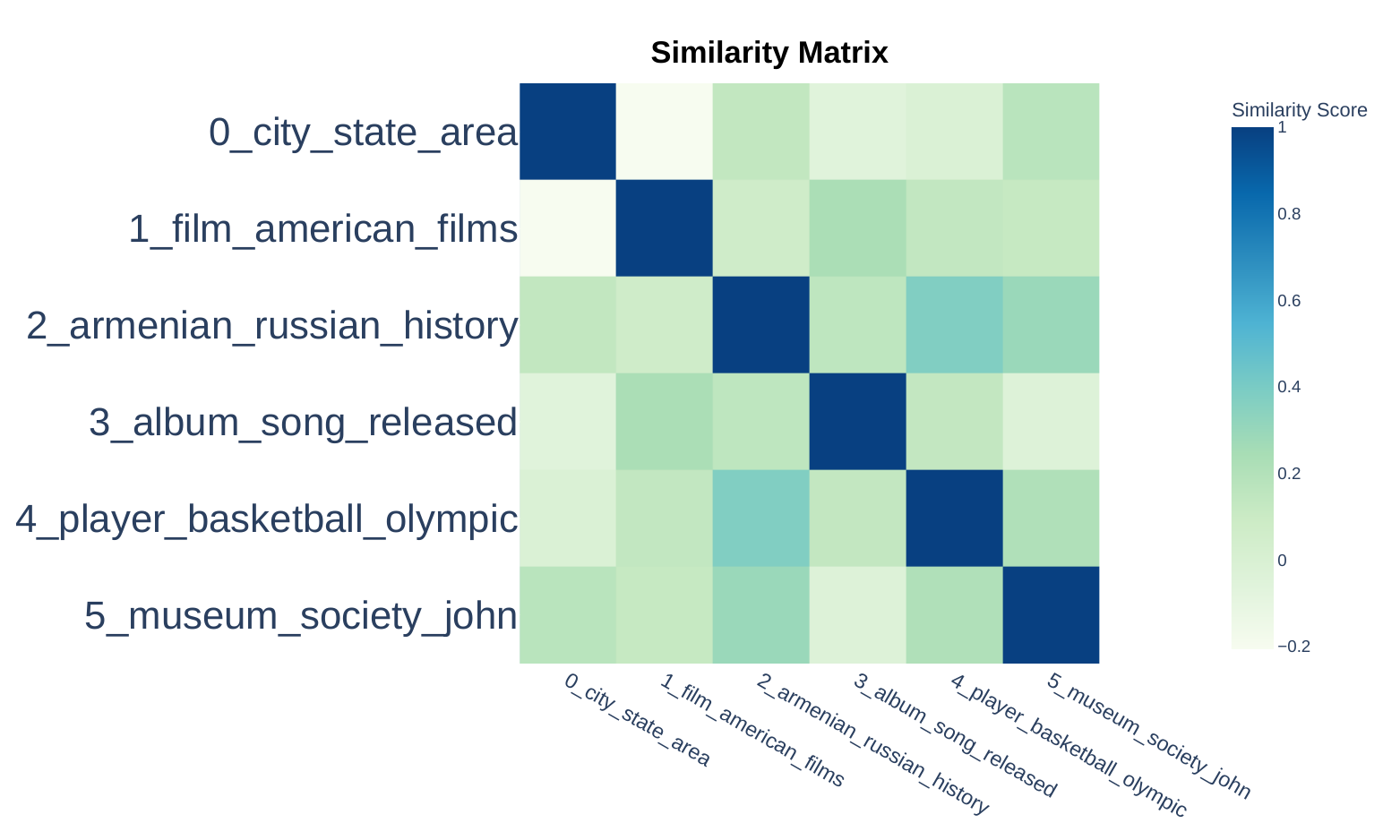}
    \caption{The similarity heatmap of the top 6 frequent topics present within the mined English paragraphs.}
    \label{fig:topic_hir}
\end{figure}

\begin{figure}[H]
    \centering
    \includegraphics[width=\columnwidth]{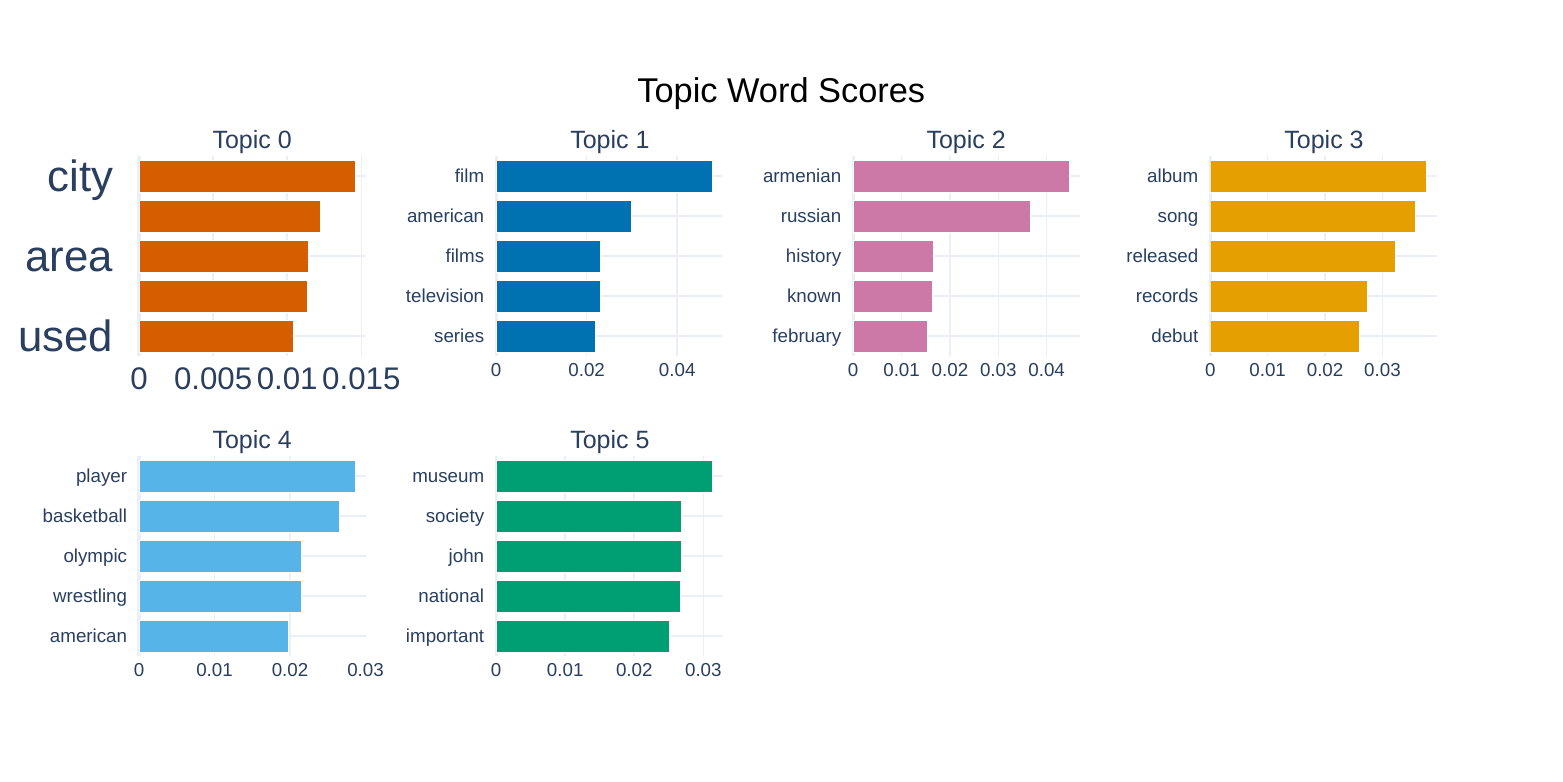}
    \caption{The usage of frequent words in the top 6 frequent topics present within the mined English paragraphs.}
    \label{fig:topic_word_distrib}
\end{figure}

\section{Appendix}
\label{sec:appendix}

\begin{table*}[hbt!]
\centering
\begin{adjustbox}{max width=\textwidth}
\begin{tabular}{lcccccccccccccccc}
\toprule
 \textbf{OTHER} & \textbf{NORP} & \textbf{GPE} & \textbf{PERCENT} & \textbf{PERSON} & \textbf{DATE} & \textbf{ORG} & \textbf{WORK OF ART} & \textbf{LANGUAGE} & \textbf{QUANTITY} & \textbf{EVENT} & \textbf{MONEY} & \textbf{LOC} & \textbf{ORDINAL} & \textbf{TIME} & \textbf{FAC} & \textbf{PRODUCT} \\
 3178 & 172 & 223 & 8 & 397 & 335 & 327 & 14 & 10 & 25 & 21 & 9 & 52 & 38 & 9 & 9 & 3 \\
\bottomrule
\end{tabular}
\end{adjustbox}
\caption{Distribution of Entities within question-answer pairs in the generated English QA dataset. The Entity labelling scheme follows \citeauthor{honnibal2020spacy}}
\label{tab:qa_ner}
\end{table*}

\paragraph{Generated Question-Answer pairs}

We showcase examples of generated and validated question-answer pairs along with their designated English paragraph $\mathcal{P}_{\text{Eng}}$ in \cref{tab:qa_gen}. These are representative samples of the generation process, further reinforced by the fact that human evaluation of the quality of the generation showed that $98\%$ of the examples are answerable and maintain quality.

\paragraph{What are the questions about ?}

To understand the type of inquiries asked within the questions, we employ a pre-trained model for Named Entity Recognition (NER) from spaCy\footnote{\url{https://spacy.io/api/entityrecognizer}} and detect all the entity types mentioned within the question-answer pairs. The results can be seen in \cref{tab:qa_ner}, showing that the object of the inquiries can vary massively from people (PERSON) and locations (LOC) to organization (ORG), numeric values (DATE, ORDINAL, TIME), etc. This further ensures that we are able to generate high-quality questions with diverse compositions and object of inquiry types.

\paragraph{Topic Distribution the parallel paragraphs}

To estimate the overlap within the topics found in the mined paragraphs, we use unsupervised topic modelling BERTopic \citep{grootendorst2022bertopic} to segment the $5$ most frequently occurring segments. We measure the overlap between these by calculating the averaged cosine distance of the topic embeddings obtained from BERTopic. The results can be seen in \cref{fig:topic_hir} and \cref{fig:topic_word_distrib}, validating our hypothesis that we are able to cover diverse themes using our parallel paragraph mining method.

\begin{table*}[ht!]
\centering
\begin{tabular}{p{\textwidth}}
\toprule
\textbf{Example 1: UEFA Champions League} \\
\midrule
Since the rebranding of the European Champion Clubs' Cup as the UEFA Champions League in 1992, 107 different players from 37 countries have scored three goals or more in a single match (a hat-trick) on 152 occasions, representing 53 clubs from 17 leagues. The first player to achieve the feat was Juul Ellerman, who scored three times for PSV Eindhoven in a 6–0 victory over Žalgiris on 16 September 1992. Lionel Messi and Cristiano Ronaldo have scored three or more goals in a match eight times each in the Champions League, more than any other player, followed by Robert Lewandowski with six, and Karim Benzema with four. \\
\textbf{Question:} What was the original name of the UEFA Champions League? \\
\textbf{Answers:} 1. European Champion Clubs' Cup, 2. European Premier League, 3. UEFA Football Cup, 4. European Soccer Championship \\
\textbf{Correct Answer:} 1. European Champion Clubs' Cup \\
\midrule
\textbf{Example 2: Sign Languages} \\
\midrule
Sign languages (also known as signed languages) are languages that use the visual-manual modality to convey meaning, instead of spoken words. Sign languages are expressed through manual articulation in combination with non-manual markers. Sign languages are full-fledged natural languages with their own grammar and lexicon. Sign languages are not universal and are usually not mutually intelligible, although there are also similarities among different sign languages. \\
\textbf{Question:} What is the primary modality used to convey meaning in sign languages? \\
\textbf{Answers:} 1. Auditory-vocal, 2. Visual-manual, 3. Tactile-kinesthetic, 4. Olfactory-gustatory \\
\textbf{Correct Answer:} 2. Visual-manual \\
\bottomrule
\end{tabular}
\caption{Examples of English paragraphs along with their generated question-answer pairs}
\label{tab:qa_gen}
\end{table*}
\begin{figure}[t!]
    \centering
    \includegraphics[width=\columnwidth]{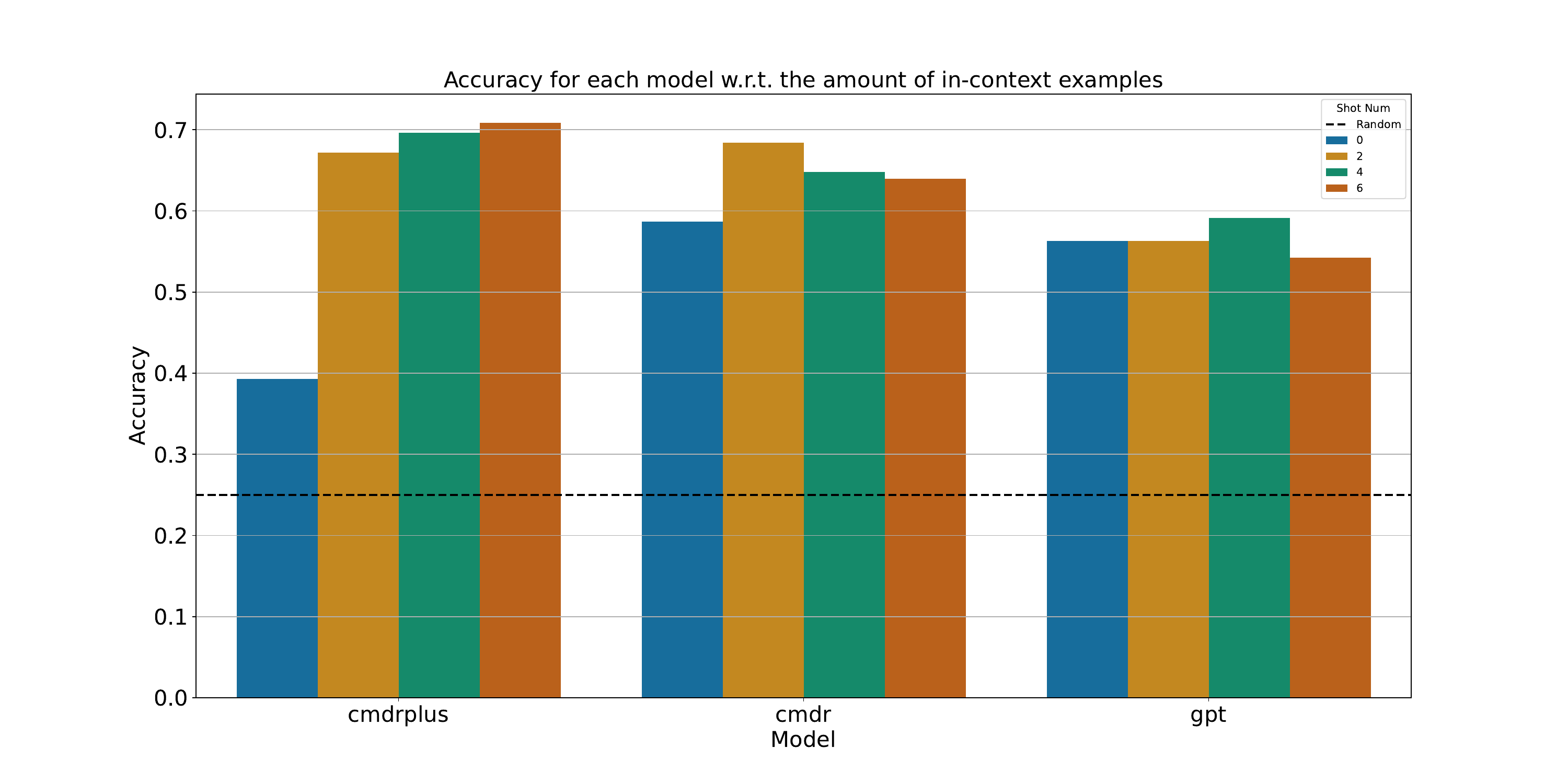}
    \caption{Accuracy of each model with a varying number of in-context examples given before generation.}
    \label{fig:topic_llm_acc}
\end{figure}

\begin{figure}[t!]
    \centering
    \includegraphics[width=\columnwidth]{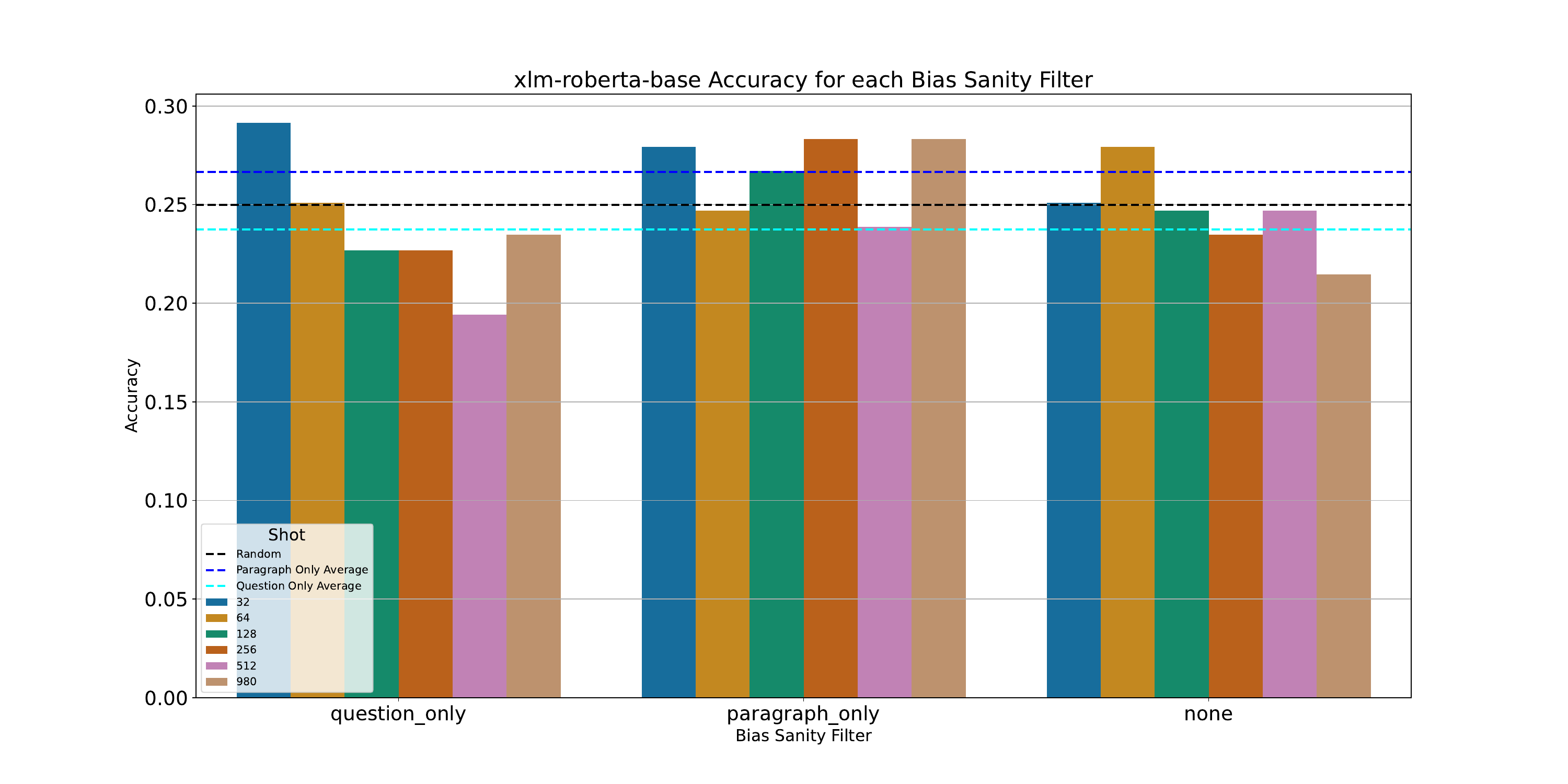}
    \caption{The results of fine-tuning XLM-Roberta on the Armenian QA dataset with a varying number of training samples while using only paragraphs, questions or random data.}
    \label{fig:topic_xlm_acc}
\end{figure}

\paragraph{Benchmarking with Armenian QA dataset}

To show the usefulness of the created dataset, we benchmark several SOTA LLMs on it in supervised fine-tuning, \emph{zero-shot} and \emph{few-shot} settings. We further investigate if the dataset suffers from statistical biases or degenerate solutions by training an XLM-RoBERTa model on inputs that contain only the paragraph or the question, excluding everything else from the sample. The results in \cref{fig:topic_xlm_acc} show us that regardless of the amount of provided training samples, the question and paragraph-only evaluations behave similarly to random chance, highlighting that the dataset is unlikely to suffer from inconsistencies and degenerate solutions.

We benchmark several LLMs, shown in \cref{fig:topic_llm_acc}, using produced Armenian QA benchmark and show that while increasing the number of model parameters and in-context samples helps the overall model performance, still even very large models are unable to solve the dataset trivially, thus showing its value as a benchmarking resource.

\end{document}